\documentclass[letterpaper, 10 pt, conference]{ieeeconf}  

\IEEEoverridecommandlockouts                              

\usepackage{epsfig} 
\usepackage{times} 
\usepackage{amsmath} 
\usepackage{amssymb}  
\usepackage{graphicx}
\usepackage{caption}
\usepackage{subcaption}
\usepackage{hyperref}
\usepackage{tikz}
\usetikzlibrary{shapes.geometric, arrows}
\usetikzlibrary{3d,fit, positioning}
\usepackage{lipsum}
\usepackage{xcolor}

\usepackage{import}
\usetikzlibrary{quotes,arrows.meta}
\usetikzlibrary{positioning}

\usepackage{Box}
\usepackage{RightBandedBox}

\graphicspath{ {./images/}}

\begin{document}

\title{\LARGE \bf Lidar Upsampling with Sliced Wasserstein Distance}

\author{Artem Savkin$^{1,2}$\\TUM, BMW \and Yida Wang$^{1}$\\TUM \and Sebastian Wirkert$^{2}$\\BMW \and Nassir Navab$^{1,3}$\\TUM, JHU \and Federico Tombari$^{1,4}$\\TUM, Google
}


\newcommand\blfootnote[1]{%
	\begingroup
	\renewcommand\thefootnote{}\footnote{#1}%
	\addtocounter{footnote}{-1}%
	\endgroup
}

\newcommand{\norm}[1]{\left\lVert#1\right\rVert}
\newcommand{\TODO}[1] {\textbf{[TODO: #1]}}
\newcommand{\Loss}{\mathcal{L}}
\newcommand{\Exp}{\mathop{\mathbb{E}}}
\newcommand{\rarr}{\rightarrow}
\newcommand{\larr}{\leftarrow}

\maketitle


\begin{abstract}
Lidar became an important component of the perception systems in autonomous driving. But challenges of training data acquisition and annotation made emphasized the role of the sensor to sensor domain adaptation. In this work, we address the problem of lidar upsampling.  Learning on lidar point clouds is rather a challenging task due to their irregular and sparse structure. Here we propose a method for lidar point cloud upsampling which can reconstruct fine-grained lidar scan patterns. The key idea is to utilize edge-aware dense convolutions for both feature extraction and feature expansion. Additionally applying a more accurate Sliced Wasserstein Distance facilitates learning of the fine lidar sweep structures. This in turn enables our method to employ a one-stage upsampling paradigm without the need for coarse and fine reconstruction. We conduct several experiments to evaluate our method and demonstrate that it provides better upsampling.
\end{abstract}

\blfootnote{
$^{1}$TUM, 85748 Munich, Germany
{\tt\small artem.savkin@tum.de};
{\tt\small tombari@in.tum.de}
}%
\blfootnote{$^{2}$BMW AG, 80809 Munich (Germany)}
\blfootnote{$^{3}$JHU, Baltimore, MD 21218, United States}
\blfootnote{$^{4}$Google, 8002 Zurich, Switzerland}


\section{Introduction}

Lidar can significantly improve the quality and reliability of the perception systems of autonomous vehicles. As an active sensor, lidar takes accurate measurements of the surrounding environment and provides spatial information about the traffic scene in form of point clouds. It is robust to challenging light conditions and can significantly contribute to the functional safety of fully autonomous systems. Also, the development and manufacturing of lidar have been democratized considerably in recent years. Democratization enabled the equipment of the serial vehicles with this sensor. As a result, lidar became a crucial component of state-of-the-art autonomous driving.

At the same time, machine perception on point clouds is still challenging since this kind of data is unstructured and irregular. However, in computer vision, neural networks have already proved superior to traditional algorithms.

\begin{figure}
\centering
\begin{minipage}{1.0\columnwidth}
\centering
\includegraphics[width=1.0\textwidth]{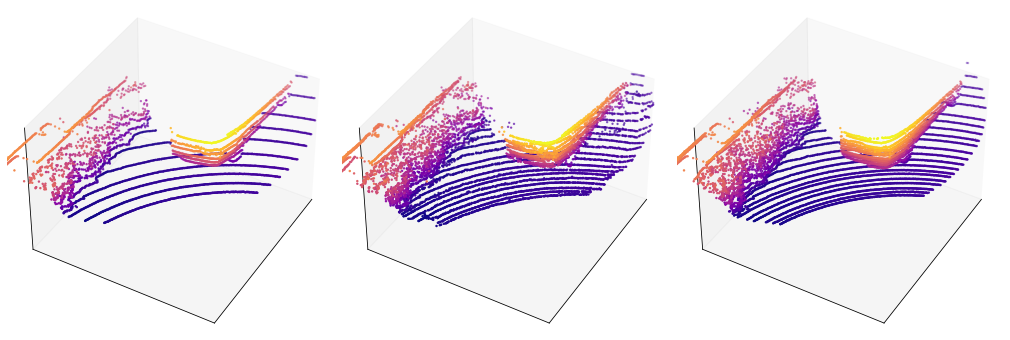}
\end{minipage}
\captionof{figure}{Results of $2 \times$ upsampling for $8129$ points patches from KITTI dataset.}
\label{fig:teaser}
\end{figure}

Their advantage is especially evident in the tasks of segmentation or detection where \textit{understanding} of the semantics of the underlying scene is involved. A wide variety of methods dedicated to 3D object detection, tracking, and segmentation successfully adapted deep learning techniques to point clouds to tackle the challenges mentioned above \cite{Qi2017, Thomas2019, Shi2019, Yin2021}.

In practical applications such as autonomous driving, data acquisition and annotation required for training such perception methods is a expensive and inflexible process.
Such rigidness is characterized by a model's performance degradation attributed to data modality change e.g., sensor viewpoint, resolution, etc. \cite{Coors2019}. For example, \cite{Besic2022} demonstrates that lidar segmentation methods reveal significant performance degradation when evaluated on the samples from a lidar in which mounting position, number of scan lines, or intensity calibration differs from the training data. Table~\ref{tab:detection} shows similar behavior, demonstrated by the object detection networks operating on lidar samples with different resolutions. The phenomenon which causes degradation is commonly referred to as domain shift \cite{Sugiyama2012} and addressed through transfer learning. They aim to address the domain shift problem by learning the domain-invariant features or by learning the direct mapping between source and target domains. Still, models are supposed to be compatible with novel sensor generations and, thus, with data modalities or domains. In this work, we address the problem of such sensor-to-sensor domain adaptation as a mechanism to avoid actual data re-acquiring or re-annotating the data. In particular, we focus here on the task with low- and high-resolution lidars. The latter is distinguished by a higher number of scan lines in the vertical field of view. Thus, the overarching goal is to sample the data points from the target distribution or, in other words, such point clouds which resemble characteristics of data samples from the target domain. Our task is to generate a point cloud sample with more scan lines given a low-resolution lidar scan sample. We consider both modalities separate domains and employ a point cloud transfer technique to solve them.

\begin{figure*}[t!]
\centering

\begin{tikzpicture}

\definecolor{color_1}{RGB}{13, 8, 135}
\definecolor{color_2}{RGB}{83,   2, 164}
\definecolor{color_3}{RGB}{140,  10, 165}
\definecolor{color_4}{RGB}{85,  51, 108}
\definecolor{color_5}{RGB}{220,  92, 104}
\definecolor{color_6}{RGB}{244, 136,  73}
\definecolor{color_7}{RGB}{254, 189,  43}
\definecolor{color_8}{RGB}{240, 249,  34}

\definecolor{plasma}{RGB}{0.050383, 0.029803, 0.527975}

\def\batchSize{1.0}

\def\abstand{0.75}

\tikzstyle{node} = [rectangle, minimum size=1cm, text centered, draw=black!50]
\tikzstyle{pcl} = [rectangle, fill=black!0]
\tikzstyle{layer} = [rectangle, minimum size=20mm, draw=black!20]
\tikzstyle{encoder} = [trapezium, text centered, draw=black]
\tikzstyle{vector} = [rectangle, minimum width=0.5cm, text centered]

\tikzstyle{every node}=[font=\scriptsize]

\node(source) [pcl, xshift=0cm]
{\includegraphics[width=20mm]{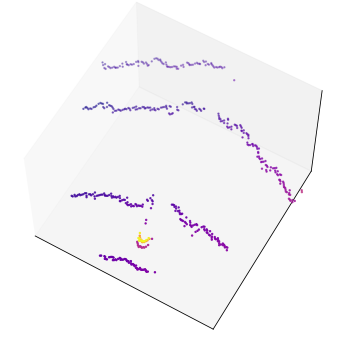}};


\node[shift={(0.0, -2.0, 0.0)}] at (source) [style={rectangle, fill=color_1, minimum size=4}] (legend1) {};
\node(text1) [right=.1 of legend1]{Conv1d};

\node[style={rectangle, fill=color_7!70, minimum size=4}] (legend2) [right of=text1] {};
\node(text2) [right=.1 of legend2] {EdgeConv};

\node[style={rectangle, fill=color_3!70, minimum size=4}] (legend3) [right of=text2] {};
\node(text3) [right=.1 of legend3] {Sample};

\node[style={rectangle, fill=color_4!70, minimum size=4}] (legend8) [right of=text3] {};
\node(text4) [right=.1 of legend8] {Linear};

\node[style={rectangle, fill=color_8!70, minimum size=4}] (legend4) [right of=text4] {};
\node(text5) [right=.1 of legend4] {Maxpool};

\node[style={rectangle, draw=gray, fill=white, minimum size=4}] (legend5) [right of=text5] {};
\node(text6) [right=.1 of legend5] {Repl.};

\node[style={rectangle, fill=color_6!70, minimum size=4}] (legend6) [right of=text6] {};
\node(text7) [right=.1 of legend6] {Interp.};

\node[style={rectangle, fill=color_2!70, minimum size=4}] (legend7) [right of=text7] {};
\node(text8) [right=.1 of legend7] {Reshape};


\pic[shift={(2.0, 0, 0)}] at (source) {Box={name=conv1d_1, caption=, fill=color_1!70, draw=red, opacity=1.0, height=8, width=\batchSize, depth=8}};

\node[shift={(2.0, -1.5, 0.0)}] at (source.west){$N \times d$};

\pic[shift={(0.0,0 , 0)}] at (conv1d_1-east) {Box={name=dense_1, caption=, fill=color_7!70, color=, opacity=1.0, height=8, width=\batchSize, depth=8}};

\pic[shift={(0.0,0 , 0)}] at (dense_1-east) {Box={name=pool_1, caption=, fill=color_3!70, color=, opacity=1.0, height=8, width=\batchSize, depth=8}};

\draw[->, draw=black!50] (source.east) -- ([xshift=-0.5cm]conv1d_1-west);

\node[shift={(0.0, -1.5, 0.0)}] at (conv1d_1-east){$N \times C$};


\pic[shift={(\abstand, 0, 0)}] at (pool_1-east) {Box={name=conv1d_2, caption=, fill=color_1!70, opacity=1.0, height=6, width=\batchSize, depth=6}};

\pic[shift={(0.0, 0, 0)}] at (conv1d_2-east) {Box={name=dense_2, caption=, fill=color_7!70, opacity=1.0, height=6, width=\batchSize, depth=6}};

\pic[shift={(0.0, 0, 0)}] at (dense_2-east) {Box={name=pool_2, caption=, fill=color_3!70, opacity=1.0, height=6, width=\batchSize, depth=6}};


\node[shift={(0.0, -1.5, 0.0)}] at (conv1d_2-east){$N/r \times C$};

\pic[shift={(\abstand, 0, 0)}] at (pool_2-east) {Box={name=max, caption=, fill=color_4!70, opacity=1.0, height=4, width=\batchSize, depth=4}};

\pic[shift={(0.0, 0, 0)}] at (max-east) {Box={name=mlp_1, caption=, fill=color_8!70, opacity=1.0, height=4, width=\batchSize, depth=4}};

\pic[shift={(\abstand, 0, 0)}] at (mlp_1-east) {Box={name=feat_3, caption=, fill=, opacity=0.0, height=1, width=1, depth=4}};

\node[shift={(0.0, -1.5, 0.0)}] at (feat_3-east){$C$};

\draw[->, draw=black!50] ([xshift=0.0cm]mlp_1-east) -- ([xshift=-0.5cm]feat_3-east);


\pic[shift={(\abstand, 0, 0)}] at (feat_3-east) {Box={name=feat_5, caption=, fill=, opacity=0.0, height=4, width=1, depth=4}};

\draw[densely dotted, draw=black!50] ([xshift=0.0cm]feat_3-east) -- ([xshift=-0.5cm]feat_5-east);


\pic[shift={(\abstand, 0, 0)}] at (feat_5-east) {Box={name=int_1, caption=, fill=color_6!70, opacity=1.0, height=6, width=\batchSize, depth=6}};

\draw[->, draw=black!50] ([xshift=0.0cm]feat_5-east) -- ([xshift=-0.5cm]int_1-east);

\pic[shift={(0.0, 0, 0)}] at (int_1-east) {Box={name=conv1d_7, caption=, fill=color_1!70, opacity=1.0, height=6, width=\batchSize, depth=6}};

\node[shift={(-0.5, -1.5, 0.0)}] at (conv1d_7-west){$N/r \times C$};

\pic[shift={(\abstand, 0, 0)}] at (conv1d_7-east) {Box={name=int_2, caption=, fill=color_6!70, opacity=1.0, height=8, width=\batchSize, depth=8}};

\pic[shift={(0.0, 0, 0)}] at (int_2-east) {Box={name=conv1d_8, caption=, fill=color_1!70, opacity=1.0, height=8, width=\batchSize, depth=8}};

\node(extractor)[shift={(-0.25, -1.5, 0.0)}] at (conv1d_8-west){$N \times C$};

\pic[shift={(\abstand, 0, 0)}] at (conv1d_8-east) {Box={name=dense_3, caption=, fill=color_7!70, opacity=1.0, height=8, width=\batchSize, depth=8}};

\node[shift={(0.0, -1.5, 0.0)}] at (dense_3-west){$N \times 2C$};

\pic[shift={(\abstand, 0, 0)}] at (dense_3-east) {Box={name=reshape, caption=, fill=color_2!70, opacity=1.0, height=8, width=\batchSize, depth=8}};

\node[shift={(0.0, -1.5, 0.0)}] at (reshape-east){$2N \times C$};

\pic[shift={(\abstand, 0, 0)}] at (reshape-east) {Box={name=mlp, caption=, fill=color_4!70, opacity=1.0, height=8, width=\batchSize, depth=8}};

\node(target) [pcl, shift={(2.0, 0, 0)}] at (mlp-east)
{\includegraphics[width=20mm]{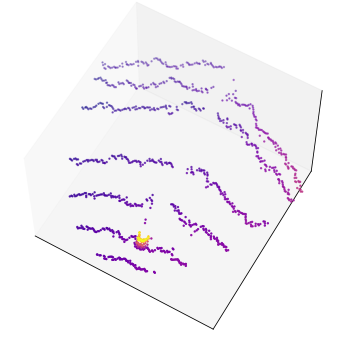}};

\node[shift={(0.0, -1.5, 0.0)}] at (target.west){$2N \times d$};

\draw[->, draw=black!50] ([xshift=0.5cm]mlp-east) -- ([xshift=0.0cm]target.west);

\draw[dotted] (source.north) -- node{} ++ (0, 0.5, 0) -- node{}  ++ (9.0, 0, 0) -- node{} ++ (0, -0.5, 0);

\draw[dotted] (9.1, 1.15, 0) -- node{} ++ (0, 0.475, 0) -- node{}  ++ (2.0, 0, 0) -- node{} ++ (0, -0.485, 0);

\draw[dotted] (11.2, 1.15, 0) -- node{} ++ (0, 0.475, 0) -- node{}  ++ (3.0, 0, 0) -- node{} ++ (0, -0.485, 0);

\node(style_a)[xshift=4.5cm,yshift=1.45cm]{$Ext$};
\node(style_a)[xshift=10.0cm,yshift=1.45cm]{$Exp$};
\node(style_a)[xshift=12.5cm,yshift=1.45cm]{$Rec$};

\end{tikzpicture}
\caption{Overview of the network with feature extraction (\textit{Ext}), feature expansion (\textit{Exp}) and coordinate reconstruction (\textit{Rec}).}\label{fig:scheme}

\end{figure*}
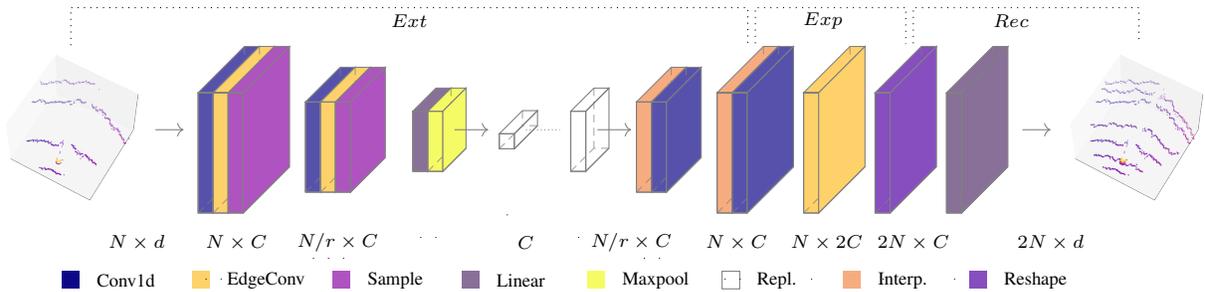

Existing methods for lidar generation mainly rely on distance measures such as Chamfer Distance (CD) and Earth Mover's Distance (EMD) \cite{Yu2018, Zhang2018}. Although several works \cite{Fan2017, Achlioptas2018} establish the superiority of EMD over CD as more discriminative for point cloud learning, the CD simply nearest neighbor based is still more efficient. Yet, from our observations, both measures do not suffice to generate fine-structure lidar point clouds. CD fails to equally distribute additional points across the lidar scan, thus erroneously favoring the known or source point locations. Standard EMD approximation based on auction algorithm \cite{Fan2017} fails to reconstruct the fine scan pattern of lidar sweeps accurately. This phenomenon is also confirmed by experiments in \cite{Yang2019}. Multiple works have addressed the computational complexity of EMD and the whole family of Wasserstein distances (WD) \cite{Li2018, Kolouri2019, Wu2019, Nguyen2021} in various tasks.

The contribution of this work is twofold. Firstly, it provides an extensive analysis of applicability of established distance metrics such as Chamfer distance, approximated EMD, and SWD for lidar scan generation and upsampling. Secondly, based on this analysis, it proposes a new lidar upsampling method, which is dedicated to the problem of cross-sensor domain adaptation between low-resolution and high-resolution lidars. We argue that, unlike existing point cloud upsampling methods, our edge-aware one-stage (no coarse reconstruction) approach based on Sliced-Wasserstein distance can reconstruct fine details of lidar scans and surpass existing upsampling techniques.

\section{Related work}

\begin{figure*}[t!]
\centering

\begin{subfigure}{0.245\textwidth}
\includegraphics[width=1.0\textwidth]{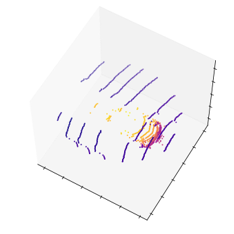}
\caption{Source}
\end{subfigure}
\begin{subfigure}{0.245\textwidth}
\includegraphics[width=1.0\textwidth]{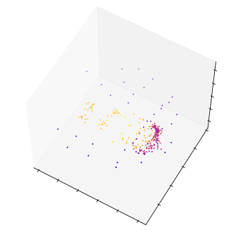}
\caption{CD}
\end{subfigure}
\begin{subfigure}{0.245\textwidth}
\includegraphics[width=1.0\textwidth]{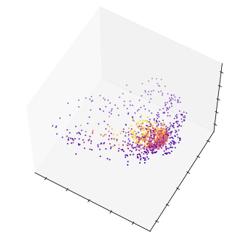}
\caption{EMD}
\end{subfigure}
\begin{subfigure}{0.245\textwidth}
\includegraphics[width=1.0\textwidth]{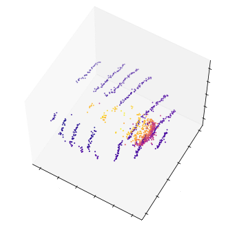}
\caption{Ours}
\end{subfigure}

\caption{Point cloud reconstruction using PointNetFCAE with conventional loss functions.}
\label{fig:metric_cd_emd_patch}
\end{figure*}
The lidar super-resolution or lidar upsampling task has been addressed by the recent learning-based methods, which can be categorized into grid-based methods \cite{Shan2020} and point-based \cite{Li2019} methods.
\subsection{Grid-based lidar generation}
Given an input condition, e.g., vector, the generation task is to produce a $N \times d$ point cloud of size $N$ and dimensionality $d$. Grid-based methods utilize the range image representation of point clouds projected into the cylindrical coordinate system. \cite{Caccia2019} adapts generative adversarial models applied on such range images to produce large-scale lidar scans. Unlike the lidar GAN approach, which uses real lidar scans for the discriminator training, \cite{Shan2020} uses training exclusively from simulated data to produce scans comparable with real high-resolution lidar. \cite{Kwon2022} improves range image upscaling by utilizing a widely acclaimed attention mechanism. Finally, \cite{Jung2022} approaches integrating an uncertainty estimation mechanism into the upsampling process to meet real-time constraints. In this work, we aim to develop a rather generic domain adaptation technique for lidar upsampling, which falls into the latter category of point-set-based methods.
\subsection{Point-based lidar generation}
 Typical point-based generation methods such as AtlasNet \cite{Groueix2018}, and FoldingNet \cite{Yang2018} operate directly on the point sets where they approach the problem by creating 2D patches and aligning them to the underlying point cloud shape. With the emergence of generative neural nets, several works like PCGAN \cite{Achlioptas2018}, GCN-GAN \cite{Valsesia2019}, and TreeGAN \cite{Shu2019} applied adversarial training frameworks to the point clouds. Luo \textit{et al.} \cite{Luo2021} employed a concept similar to image diffusion for the point cloud generation. More recent generative methods utilize normalizing flows - PointFlow \cite{Yang2019}, gradient fields - ShapeGF \cite{Cai2020} and autoregression - PointGrow \cite{Sun2020}.
 
Compared to generation, the upsampling aims to compute a $rN \times d$ output given a $N \times d$ point cloud, with upsampling ratio $r$ so that it resembles the spatial characteristics of the underlying 3d shape.
Methods introduced by Yu \textit{et al.} \cite{Yu2018PuNet} and Zhang \textit{et al.} \cite{Zhang2018} were pioneering deep learning-based approaches for that task. PU-Net \cite{Yu2018PuNet} relied on Point-Net++ encoder \cite{Qi2017} for feature extraction from patches then applied two sets of linear layers onto features replicated from $N \times C$ to $N \times rC$ which are then merely reshaped into $rN \times C$. PU-Net is followed by EC-net \cite{Yu2018}, which introduced a loss function that aimed to minimize the point-to-edge distance to make the network more edge-aware. Later 3PU \cite{Yifan2019} explored multi-scale progressive patch-based upsampling by breaking down a $r$-factor upsampling network into several $2 \times$ upsampling networks. It introduces dense connections or blocks for the feature extraction and defines local neighborhoods later in a feature space via kNN based on feature similarity. PU-GAN \cite{Li2019} adopts an adversarial framework for point cloud upsampling by formulating a discriminator network that learns to evaluate generated high-resolution point clouds. It introduces a \textit{up-down-up} expansion unit, which enables a so-called self-correction of expanded features. More recent methods include PU-GCN \cite{Qian2021} and Dis-PU\cite{Li2021}.

Another task worth mentioning is point cloud completion, whose goal is to generate complete point clouds and create dense scans from incomplete sparse ones. Here pioneering learning method was PCN \cite{Yuan2018}. PCN first learns the global feature of the underlying shape from partial observation and then reconstructs a complete but coarse point cloud of lower resolution, which is up-sampled with folding operations in the second phase. TopNet \cite{Tchapmi2019} method utilizes a tree structure for generating complete dense point clouds. Following the two-stage coarse-fine point cloud generation MSN \cite{Liu2020} uses the morphing principle and joint CD and EMD loss. Two-staged ECG \cite{Pan2020} improves on the generation of the edges by exploiting the edge-preserved pooling. Finally, CDN \cite{Wang2020} extends a traditional GAN framework to guarantee that local areas follow the same pattern as ground truth.

Compared to previous point cloud generation works, our method is dedicated to precisely reconstructing fine-grained lidar point clouds. Discussed point-based methods reveal a relatively restricted applicability area that embraces simplistic, ShapeNet-alike \cite{Chang2015} data. Point clouds generated by a lidar sensor demonstrate significantly more complex underlying geometry, scan pattern, and data variance. Typically methods trained on PCN dataset \cite{Yuan2018} derived from synthetic ShapeNet CAD models embrace eight categories of objects (e.g., plane, car, chair) whereas lidar scans of the KITTI dataset \cite{Geiger2012} in turn were obtained during drives in the real world and contain multiple classes of objects in a single scan. Thus, the task of point cloud generation in the lidar domain remains under-explored. To our best knowledge, the proposed method is the only point-based method able to reconstruct complex lidar point clouds in the upsampling task accurately.
\section{Approach}

Our method relies on the edge-aware feature extraction, and extension mechanism \cite{Yifan2019} and projection-based 1-Wasserstein distance. This training objective reveals significantly improved sensitivity to fine-grained lidar scan patterns as opposed to conventional loss functions.

\subsection{Model}

\begin{figure*}[t!]
\centering

\begin{subfigure}{0.195\textwidth}
\includegraphics[width=1\textwidth]{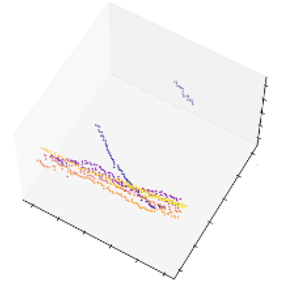}
\includegraphics[width=1\textwidth]{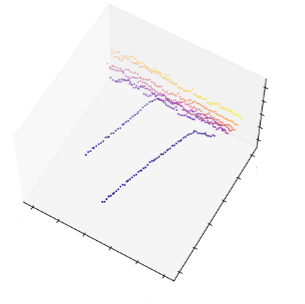}
\caption{Source}
\end{subfigure}
\begin{subfigure}{0.195\textwidth}
\includegraphics[width=1\textwidth]{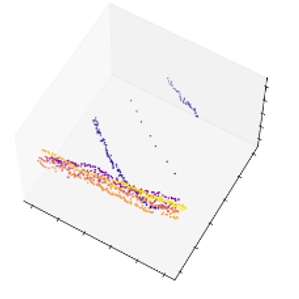}
\includegraphics[width=1\textwidth]{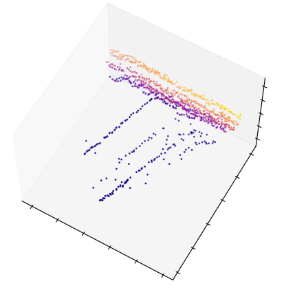}
\caption{CD}
\end{subfigure}
\begin{subfigure}{0.195\textwidth}
\includegraphics[width=1\textwidth]{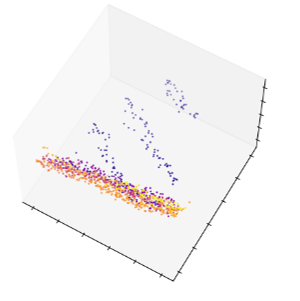}
\includegraphics[width=1\textwidth]{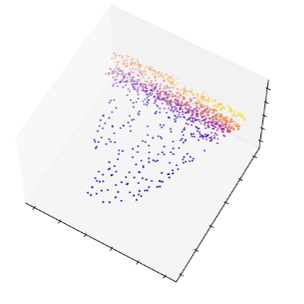}
\caption{EMD}
\end{subfigure}
\begin{subfigure}{0.195\textwidth}
\includegraphics[width=1\textwidth]{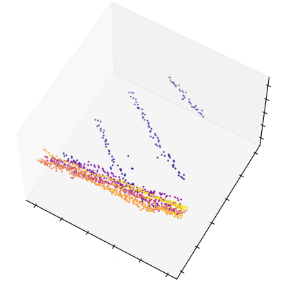}
\includegraphics[width=1\textwidth]{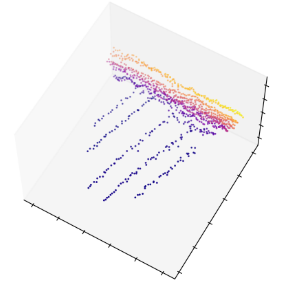}
\caption{SWD}
\end{subfigure}
\begin{subfigure}{0.195\textwidth}
\includegraphics[width=1\textwidth]{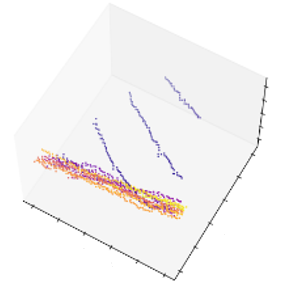}
\includegraphics[width=1\textwidth]{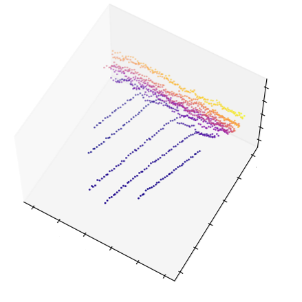}
\caption{Ground Truth}
\end{subfigure}
\caption{Lidar upsampling using proposed network with conventional loss functions.}
\label{fig:ablation_study}
\end{figure*}

The proposed model is built upon established upsampling carcass with three steps: feature extraction, feature expansion, and coordinate reconstruction. The feature extractor is implemented as a plain encoder-decoder architecture and similarly to 3PU ~\cite{Yifan2019} employs dense per point feature learning using \textit{Conv1d} and \textit{EdgeConv} \cite{Wang2019} layers. An \textit{EdgeConv} layer applies the kNN algorithm to find clusters of points and preserves characteristic attributes like edges. Following the edge-preserving principle, the feature expansion module also takes advantage of \textit{EdgeConv} layer for generating the $N \times 2C$ upsampled feature vector. In the final step, a set of linear layers reconstruct the resulting $N \times 3$ coordinates.

\begin{figure}[t!]
\centering
\includegraphics[width=.95\columnwidth]{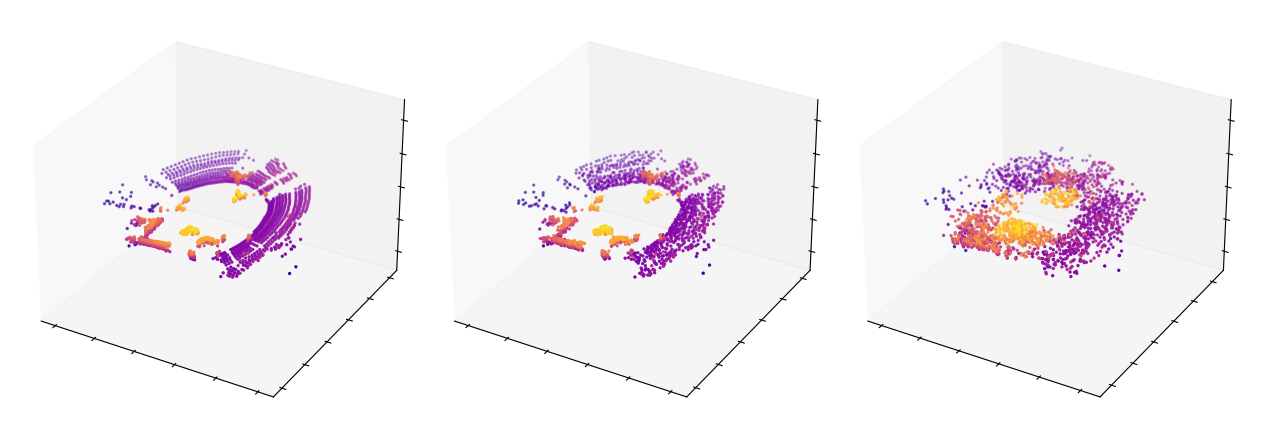}
\includegraphics[width=.95\columnwidth]{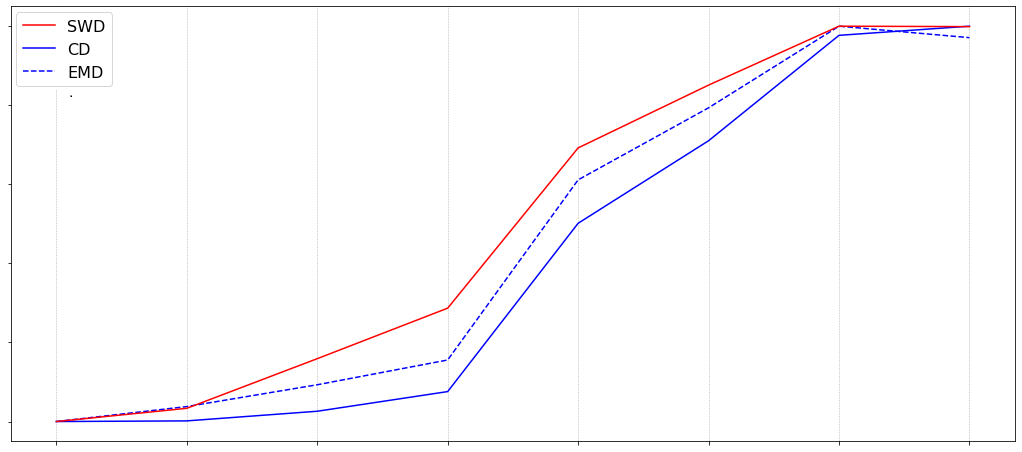}
\caption{Distance metrics on jitter (normalized).}
\label{fig:metric_jitter_plot}
\end{figure}
\begin{figure}[t!]
\centering
\includegraphics[width=.95\columnwidth]{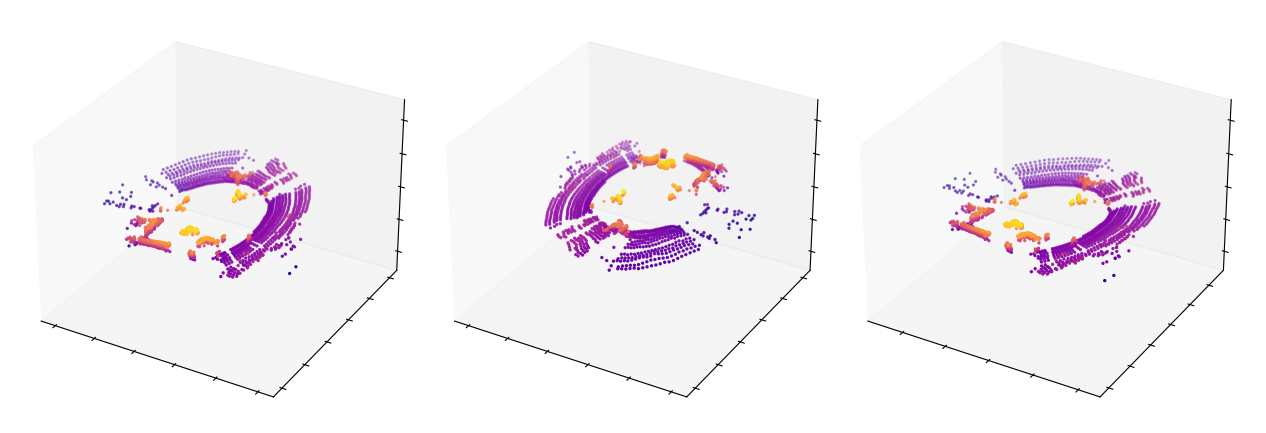}
\includegraphics[width=.95\columnwidth]{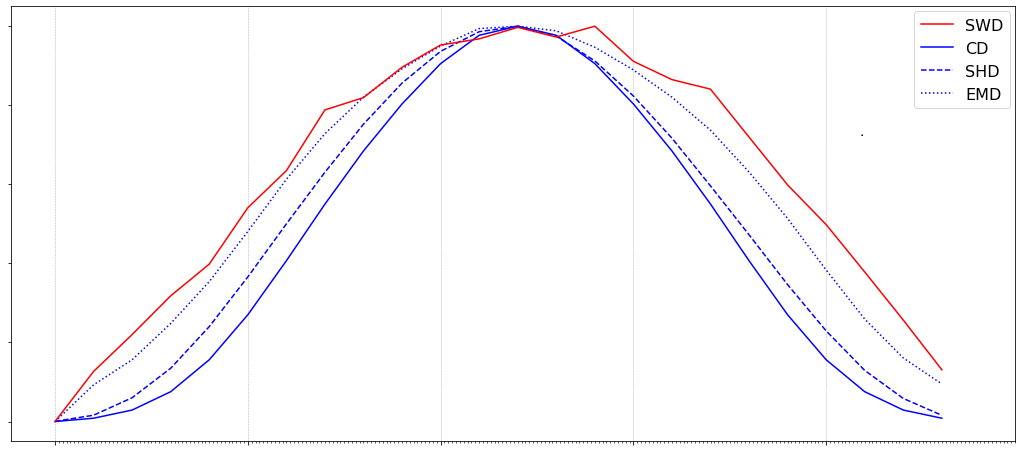}
\caption{Distance metrics on rotation (normalized).}
\label{fig:metric_rotation_plot}
\end{figure}

\textit{Feature Extraction}. As mentioned before, the encoder first takes the $N \times d$ point cloud as input and translates it into $N/r \times C'$ feature vector by applying several $Conv1d$ plus $EdgeConv$ blocks. Here, $N$ is the original number of points, $r$ is downsampling ratio, $d$ - point cloud dimensionality (in our experiments $d = 3$) and $C'$ is the resulting size of the feature vector. The common farthest point sampling technique achieves downsampling from $N$ to $N/r$. This feature vector is transformed to $C'$-sized global feature, which will be upscaled into $N \times C$ by replication and interpolation. For that, replicated $C'$ feature vectors are just stacked together and interpolated using inverse distance weighted average from \cite{Qi2017}.

\textit{Feature Expansion}. Inspired by \cite{Pan2020}, we design the feature expansion module along the lines of edge awareness by utilizing the \textit{EdgeConv} layer. This layer employs the kNN algorithm and enables the module to find clusters based on similarities in the feature space. It takes a $N \times C$ learned feature vector and transforms it into $N \times 2C$ upsampled feature vector and then reshaped to $2N \times C$ tensor.

\textit{Coordinates Reconstruction.}  the resulting tensor will be then processed by a set of linear layers to reconstruct the point dimensions $2N \times d$. Fig.~\ref{fig:scheme} visualizes the overall architecture.

\subsection{Loss Function}

We argue that conventional loss functions based on Chamfer Distance (CD) and approximations of Earth Mover's Distance (EMD) \cite{Liu2020} suffer from the lack of \textit{sensitivity} to the fine-grained scan patterns of lidar sensors. As mentioned before, CD unequally distributes generated points across the lidar scan in point cloud reconstruction and favors the known or source point locations. A possible reason for that could be the nature of a generative task where produced point clouds at the initial state represent unstructured point blobs so that CD fails to find correspondences. On the other hand, the reason for EMD to fail inaccurate reconstruction of the fine scan patterns could lie in commonly used approximations which lead to biased or noisy gradients \cite{Yang2019} and herewith affect the ability of a model to distinguish the fine details.

\begin{figure*}[t!]

\begin{subfigure}{0.16\textwidth}
\includegraphics[width=1\textwidth]{images/skitti/input/104.png}
\includegraphics[width=1\textwidth]{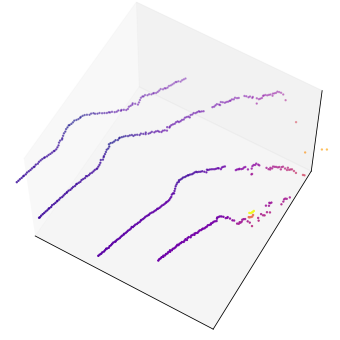}
\includegraphics[width=1\textwidth]{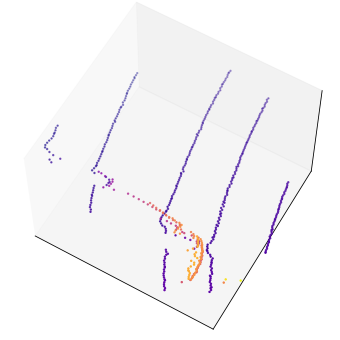}
\caption{Source}
\end{subfigure}
\begin{subfigure}{0.16\textwidth}
\includegraphics[width=1\textwidth]{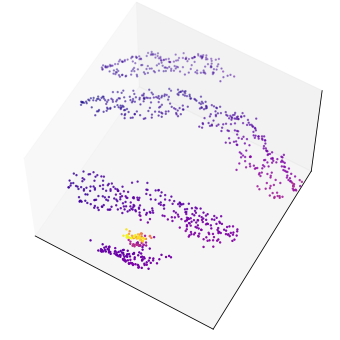}
\includegraphics[width=1\textwidth]{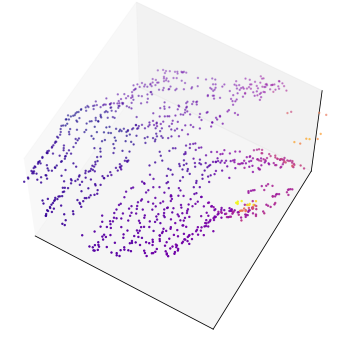}
\includegraphics[width=1\textwidth]{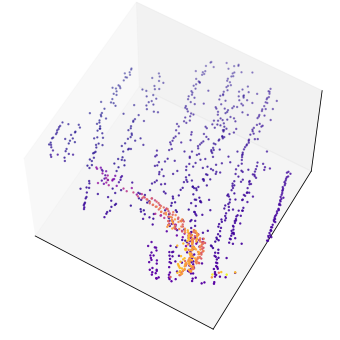}
\caption{PU-Net}
\end{subfigure}
\begin{subfigure}{0.16\textwidth}
\includegraphics[width=1\textwidth]{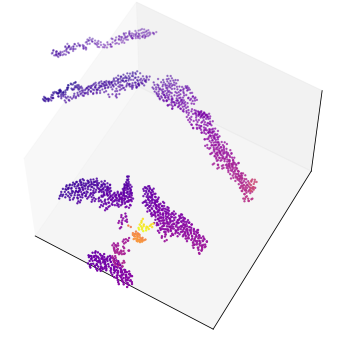}
\includegraphics[width=1\textwidth]{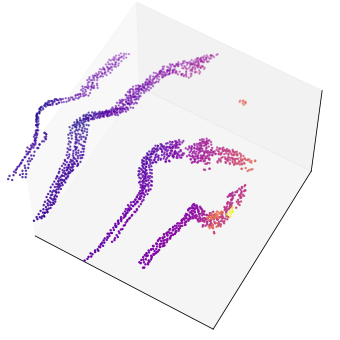}
\includegraphics[width=1\textwidth]{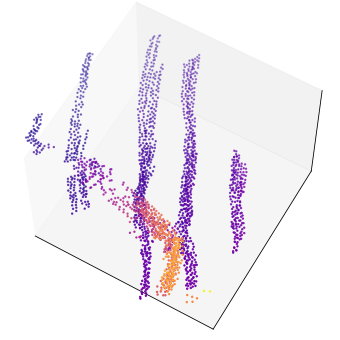}
\caption{3PU}
\end{subfigure}
\begin{subfigure}{0.16\textwidth}
\includegraphics[width=1\textwidth]{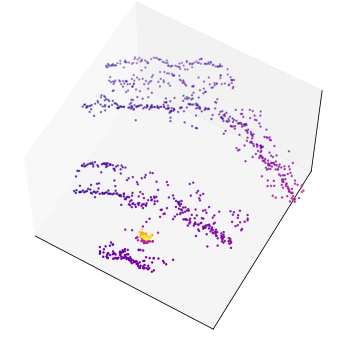}
\includegraphics[width=1\textwidth]{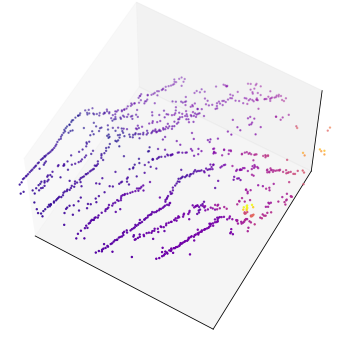}
\includegraphics[width=1\textwidth]{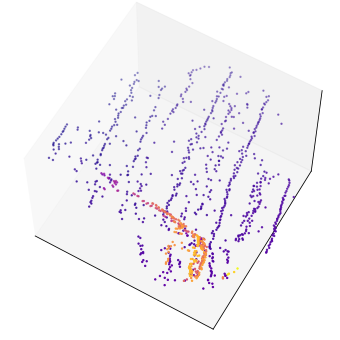}
\caption{PU-GAN}
\end{subfigure}
\begin{subfigure}{0.16\textwidth}
\includegraphics[width=1\textwidth]{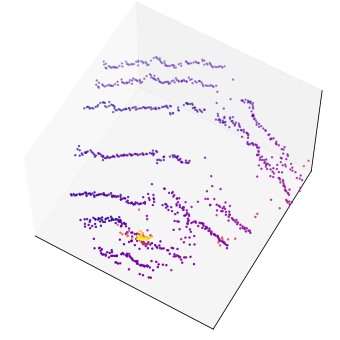}
\includegraphics[width=1\textwidth]{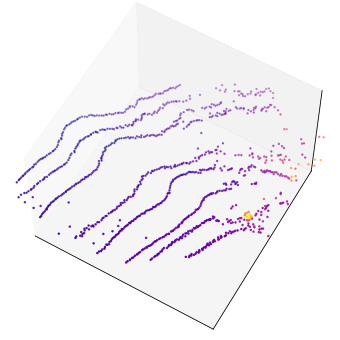}
\includegraphics[width=1\textwidth]{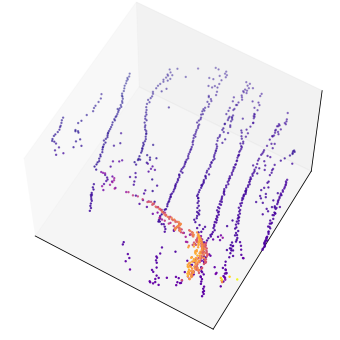}
\caption{Ours}
\end{subfigure}
\begin{subfigure}{0.16\textwidth}
\includegraphics[width=1\textwidth]{images/skitti/gt/104.png}
\includegraphics[width=1\textwidth]{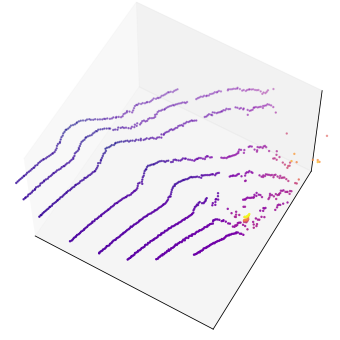}
\includegraphics[width=1\textwidth]{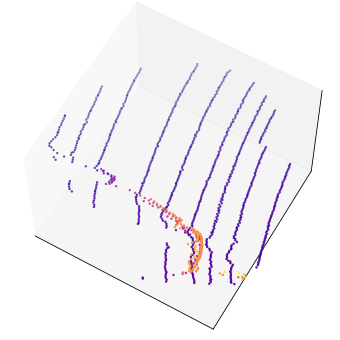}
\caption{Ground truth}
\end{subfigure}

\caption{Qualitative comparison of the $2 \times$ lidar upsampling results on KITTI \textit{test}.}
\label{fig:qualitative_results}
\end{figure*}

Following the Optimal Transport (OT) perspective, we consider the distance between point clouds as the distance between two distributions with a natural way to measure it by a metric function defined between probability distributions - Wasserstein distance. Yet accurate, exact computation of this metric is computationally intractable. The assumption is that more accurate distance calculation is a key to precise lidar scan reconstruction and upsampling. We argue that higher accuracy of point cloud reconstruction, which is sensitive to fine scan patterns, can be achieved by applying the Sliced Wasserstein (SW) distance as it possesses equivalent statistical properties to WD \cite{Kolouri2019, NguyenTrung2021}. The intuition behind SW distance resides in the idea of the existence of a closed-form OT solution in a uni-dimensional space. For two point sets $X$ and $Y$ of equal size such as $X, Y \subset \mathbb{R}^3: |X| = |Y|$, EMD between $X$ and $Y$ can be calculated as:
\begin{equation}
    EMD(X,Y) = \min_{\phi : X \rightarrow Y} \sum_{x \in X} \|x - \phi(x)\|_2
    \label{eq:emd}
\end{equation}
where $\phi: X \rightarrow Y$ is a bijection \cite{Fan2017}. The more general form of EMD in the case of $|X|=|Y|$:

\begin{equation}
    EMD(X, Y) = \inf_{P \in \Pi(X, Y)} \mathbb{E}_{(x,y)\sim P} [d(x,y)]
\end{equation}

is equivalent to Wasserstein distance with $p = 1$:

\begin{equation}
    W_p(X, Y) = \inf_{P \in \Pi(X, Y)} \mathbb{E}_{(x,y)\sim P} [d^p(x,y)]^{1/p}
    \label{eq:wasserstein}
\end{equation}
where $d$ is a metric and $\Pi(X,Y)$ is a set of all joint distributions of $P (x,y)$ of random variables $x$ and $y$ with marginal distributions $X$ and $Y$: $$\Pi(X,Y) = \{P_{X,Y} (x,y)\}$$.
In practice, computation of~\ref{eq:emd} or~\ref{eq:wasserstein} is intractable, so an approximation scheme such as an auction algorithm must be applied \cite{Liu2020} yet it cannot guarantee the required bijection assignment. As opposed to such approximation, the Sliced Wasserstein approach exploits the fact that $W_p$ has a closed-form solution for univariate probability distributions \cite{Kolouri2019}. Thus several works utilize this fact to calculate $W_p$ \cite{Wu2019, Deshpande2019}:

\begin{equation}
    SW_p(X, Y) = \int_{\mathbb{S}^{d-1}} W_p(X^{s}, Y^{s})ds
\end{equation}

where $X^s$ and $Y^s$ are projections of $X$ and $Y$ onto the direction $s$ of all possible directions of a unit sphere $\mathbb{S}^{d-1}$. In practice, $SW_p$ is approximated by only taking a finite number of random directions from $\mathbb{S}^{d-1}$. Moreover, in a uni-dimensional case, $d$ can be obtained by just sorting \cite{Deshpande2019}. Hence our final objective can be defined as follows:

\begin{equation}
    \Tilde{SW}_p (X,Y) = \frac{1}{|\Tilde{\mathbb{S}}|}\sum_{s \subset \mathbb{\Tilde{S}}} \frac{1}{|X|} \sum_{i} \|X^s_i - Y^s_i\|_2
\end{equation}

Thus, the final optimization problem can be formulated as follows:

\begin{equation}
    \underset{\theta}{\mathrm{argmin}} \: \mathbb{E}_{X \sim D_X} (\Tilde{SW}_p (X, f_\theta(X)))
\end{equation}

\section{Experiments}
\subsection{Datasets}
Our experiments are based on the popular public datasets SemanticKITTI~\cite{Behley2019} and Waymo Open Dataset~\cite{Waymo2020}. SemanticKITTI is a large-scale computer vision dataset that provides 22 sequences of lidar scans with overall around $43000$ scans alongside point cloud segmentation labels. The dataset provides measurements in German inner cities, residential areas, and highway traffic scenes during various daylight and weather conditions. In our experiments we operate with original KITTI~\cite{Geiger2012} dataset split of the sequences divided into \textit{train} and \textit{test} sets of the sizes $7481$ and $7518$ scans respectively. Semantic labels of the dataset contain information about $28$ classes such as \textit{road, sidewalk, person, car}. The Waymo dataset, in turn, provides $1150$ sequences of $20$ seconds each. In total, it offers around $230K$ annotated samples with the ground truth for object detection and tracking tasks. For our purposes, every $10$th lidar scan has been sampled from all data sequences, resulting in $15947$ samples training dataset and $4038$ validation dataset. Also, from the five lidars, we only use the scans provided by the top-mounted one.

\begin{figure*}[t!]
\centering

\begin{subfigure}{0.195\textwidth}
\includegraphics[width=1\textwidth]{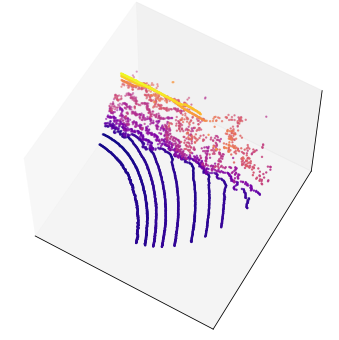}
\includegraphics[width=1\textwidth]{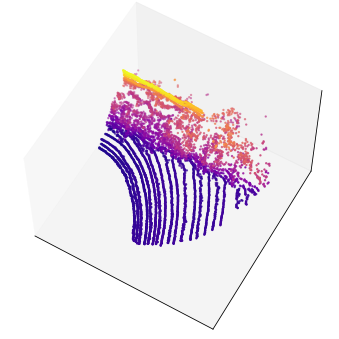}
\includegraphics[width=1\textwidth]{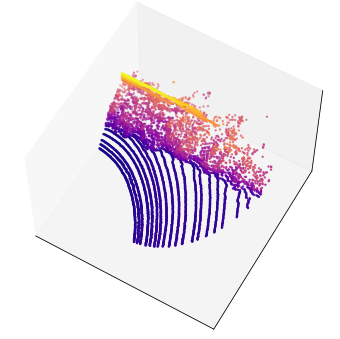}
\end{subfigure}
\begin{subfigure}{0.195\textwidth}
\includegraphics[width=1\textwidth]{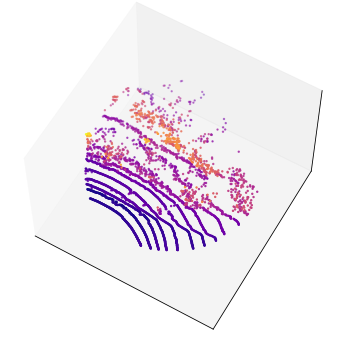}
\includegraphics[width=1\textwidth]{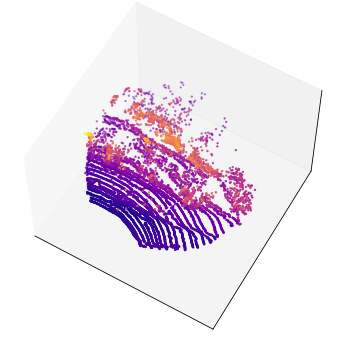}
\includegraphics[width=1\textwidth]{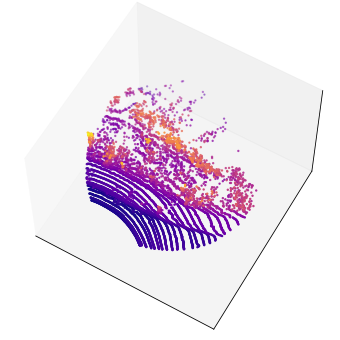}
\end{subfigure}
\begin{subfigure}{0.195\textwidth}
\includegraphics[width=1\textwidth]{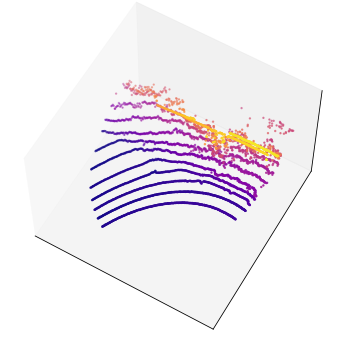}
\includegraphics[width=1\textwidth]{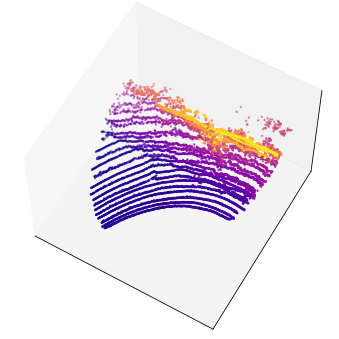}
\includegraphics[width=1\textwidth]{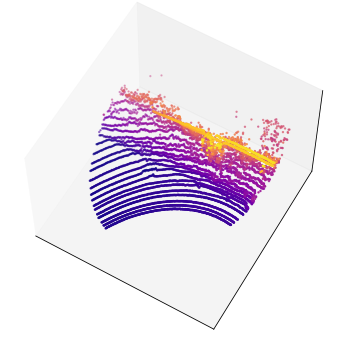}
\end{subfigure}
\begin{subfigure}{0.195\textwidth}
\includegraphics[width=1\textwidth]{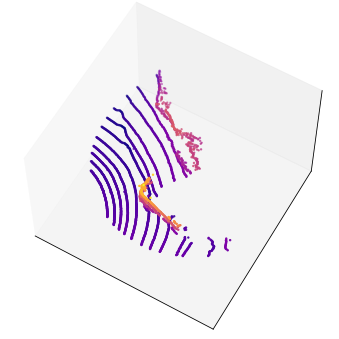}
\includegraphics[width=1\textwidth]{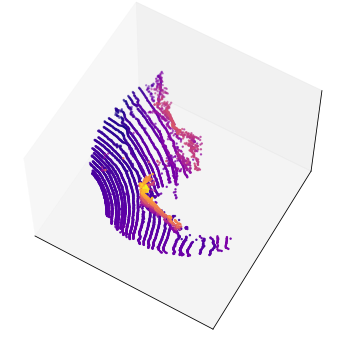}
\includegraphics[width=1\textwidth]{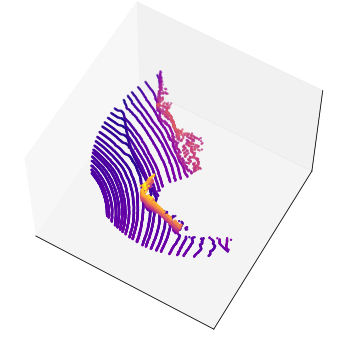}
\end{subfigure}
\begin{subfigure}{0.195\textwidth}
\includegraphics[width=1\textwidth]{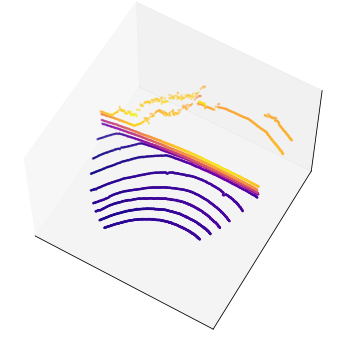}
\includegraphics[width=1\textwidth]{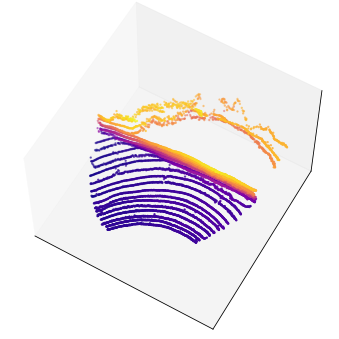}
\includegraphics[width=1\textwidth]{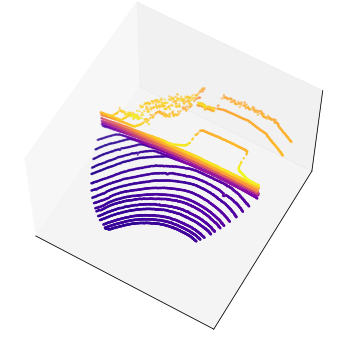}
\end{subfigure}

\caption{Lidar upsampling results (middle) on KITTI \textit{val}  patches with input samples (top) and ground truth (bottom).}
\label{fig:qualitative_results_kitti_8192pts}
\end{figure*}

\subsection{Loss} We propose to train lidar generation models using loss function based on Sliced-Wasserstein distance as opposed to commonly utilized CD, EMD-based losses. Our benchmark experiments suggest that these metrics are sufficient for a model to learn the shapes of point clouds but not the fine structures. Described behavior can be observed in Fig.~\ref{fig:metric_cd_emd_patch}, which shows the results of the reconstruction of the point cloud patches provided by a \textit{vanilla auto-encoder} network PointNetFCAE trained respectively with CD and EMD loss. This auto-encoder model consists of a Pointnet \cite{Qi2016} based encoder and a fully connected decoder.
Even though the sizes of the visualized patches are equal ($2048$ points), CD looks underpopulated in certain regions. \cite{Achlioptas2018} refers to this phenomenon as \textit{CD blindness}, which manifests in point placement where the average mass locates and fails to distinguish such poor allocation from the true one. EMD, in turn, appears to reconstruct rough point distribution but is considerably distorted. That finding is aligned with observations in \cite{Fan2017} and mirrors its mean-shape experiments.

Additionally, we compare the proposed loss function with existing error metrics concerning their \textit{sensitivity}. The results of this analysis are shown in Fig.~\ref{fig:metric_jitter_plot} and Fig.~\ref{fig:metric_rotation_plot}. The comparison is conducted between Sinkhorn loss \cite{Cuturi2013}, Sliced Gromov-Wasserstein \cite{Vayer2020}, Sliced Wasserstein Distance \cite{Kolouri2019}, Chamfer loss, and EMD loss. Here we execute two experiments by gradually applying a corresponding transformation on the point cloud and measuring the distances between transformed and original samples. In the first experiment, Gaussian noise is introduced into a sample, whereas during the second experiment, the sample is rotated along the vertical axis. In both cases, analyzed metrics behaved consistently by changing to non-zero values as the transformation became more significant in the first experiment and turned back to zero after full rotation in the second experiment. Despite that, Sliced Wasserstein Distance could identify the slightest point cloud changes by indicating higher discrepancy values in the early transformation iterations. The influence of different loss functions CD, EMD, and SWD on the lidar upsampling performed by our model will be discussed in the ablation study in the section~\ref{sec:upsampling}

\begin{table}[b]
\begin{center}
\resizebox{\columnwidth}{!}{
\begin{tabular}{l|cccc}
\hline
& CD & HD & EMD & SWD \\
\hline
PU-Net \cite{Yu2018} & 0.5272 & 1.2627 & 0.3851 & 15.4002 \\
3PU \cite{Yifan2019} & 0.1172 &0.4151 & 0.2740 & 5.2078 \\
PU-GAN \cite{Li2021} & \textbf{0.0426} & 0.1892 & 0.2504 & 1.3553 \\
Ours  & 0.0435 & \textbf{0.1694} & \textbf{0.0775} & \textbf{1.1742}  \\
\hline
\end{tabular}}    
\end{center}
\caption{Quantitative comparison for $2 \times$ upsampling of $8192$ points on KITTI \textit{val} dataset.}
\label{tab:results}
\end{table}

\begin{figure*}[t!]
\centering

\begin{subfigure}{0.195\textwidth}
\includegraphics[width=1\textwidth]{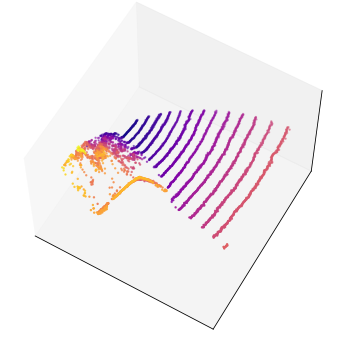}
\includegraphics[width=1\textwidth]{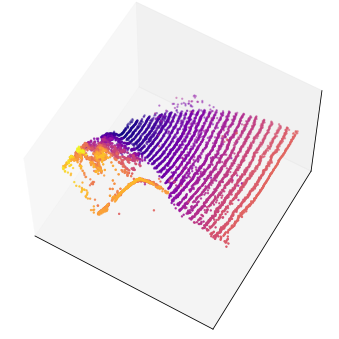}
\includegraphics[width=1\textwidth]{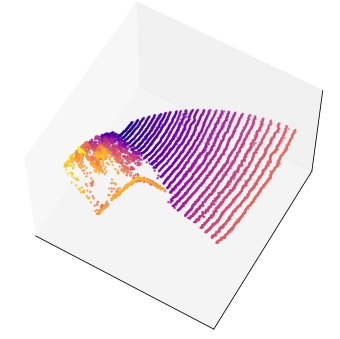}
\end{subfigure}
\begin{subfigure}{0.195\textwidth}
\includegraphics[width=1\textwidth]{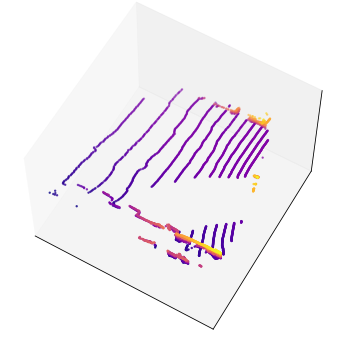}
\includegraphics[width=1\textwidth]{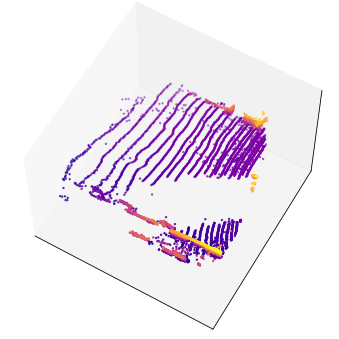}
\includegraphics[width=1\textwidth]{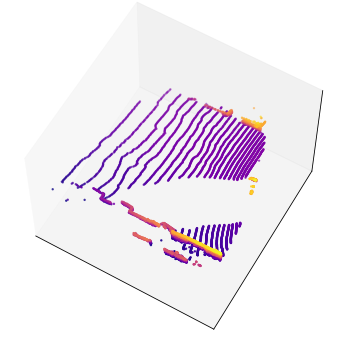}
\end{subfigure}
\begin{subfigure}{0.195\textwidth}
\includegraphics[width=1\textwidth]{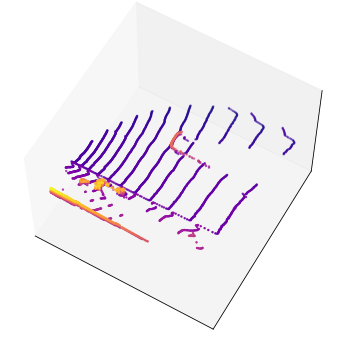}
\includegraphics[width=1\textwidth]{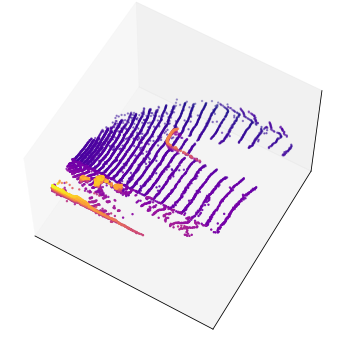}
\includegraphics[width=1\textwidth]{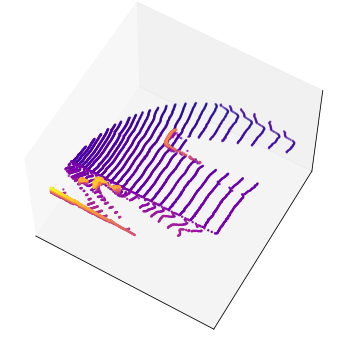}
\end{subfigure}
\begin{subfigure}{0.195\textwidth}
\includegraphics[width=1\textwidth]{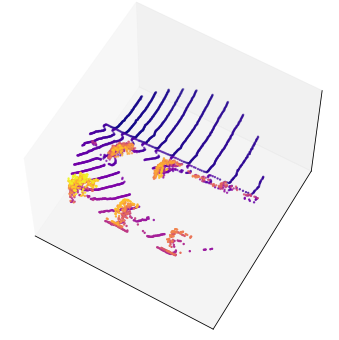}
\includegraphics[width=1\textwidth]{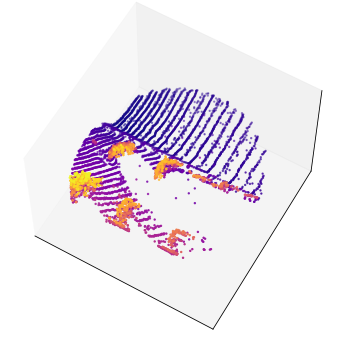}
\includegraphics[width=1\textwidth]{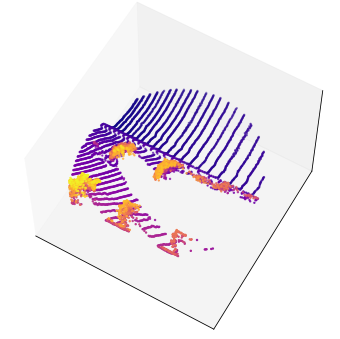}
\end{subfigure}
\begin{subfigure}{0.195\textwidth}
\includegraphics[width=1\textwidth]{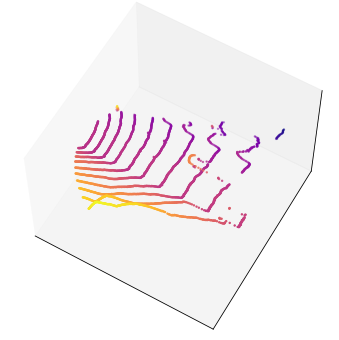}
\includegraphics[width=1\textwidth]{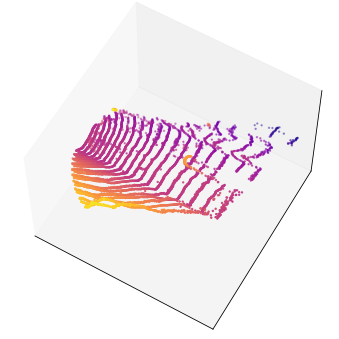}
\includegraphics[width=1\textwidth]{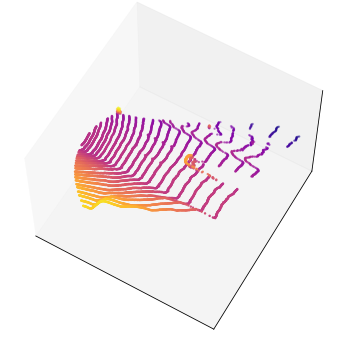}
\end{subfigure}

\caption{Lidar upsampling results (middle) on Waymo \textit{val} patches with input samples (top) and ground truth (bottom)}
\label{fig:qualitative_results_waymo_8192pts}
\end{figure*}

\subsection{Upsampling}\label{sec:upsampling}
The training occurs supervised where input patches of size $2048$ were obtained by decimating the ground truth samples by a factor of two. Decimation is performed based on rasterized input point clouds. For the Waymo dataset, every second scan line of the original cylindrical projection of size $W \times H \times 3$ of the lidar scans was dropped/rejected before projecting the points into the lidar coordinate system. Similarly, we processed the cylindrical projections of the KITTI dataset. As the KITTI dataset provides only raw point clouds, they are required to be rasterized before decimation. Therefore, KITTI scans were arranged into $64 \times  2048$ grid based on sensor vertical resolution and azimuth angle for rasterization. This preprocessing is shown in Fig.~\ref{fig:teaser}. With that data, the network is trained for $2000$ consecutive epochs without pre-trained weights initialization and using Adam optimizer. The result after the training memory footprint is $22$ Mb.
Training is accomplished on NVIDIA Quadro RTX 6000 GPU($24$GB) with a single forward pass that takes, on average $0.2$ sec. To evaluate the generated results, we conducted several experiments and assessed our results both qualitatively and quantitatively. The results of the lidar scan upsampling on the KITTI dataset are shown in Fig.~\ref{fig:qualitative_results}. To visualize the effects introduced by our approach, we display both sources and upsampled scans together with point clouds generated by the state-of-the-art upsampling methods 3PU, PU-Net, and PU-GAN. We employ original training protocols from respective reports to train those models on lidar point clouds. As reported in the original publication, PU-GAN was trained with adversarial and EMD-based reconstruction loss, 3PU was trained with modified counter-outlier CD loss, and PU-Net used plain EMD loss with repulsion loss. First, we want to draw attention to the fact that the proposed method recovers the scan lines missing in the input samples and, in consequence, effectively reconstructs the high-resolution target scans.
In comparison with the results of the existing methods, samples provided by our approach reveal no point scattering in the local neighborhood but rather provide distinguished scan lines. Another characteristic of the generated scans worth considering is the actual shape of the scans. In the upsampled lidar sweeps, recovered scan lines indeed follow the spatial characteristics of the underlying geometry.

Additionally, we evaluate our method of upsampling the point cloud patches on a larger scale. The results of $2 \times$ upsampling for patches of size $8192$ for both datasets KITTI and Waymo are depicted in Fig.~\ref{fig:qualitative_results_kitti_8192pts} and~~\ref{fig:qualitative_results_waymo_8192pts}. Here, we also want to emphasize the reconstruction quality of the upsampled scans.  

\begin{table}[b]
\begin{center}
\resizebox{\columnwidth}{!}{
\begin{tabular}{l|ccccccccc}
\hline
\rotatebox[origin=c]{0}{ Validation }
& \rotatebox[origin=c]{90}{ Car easy }
& \rotatebox[origin=c]{90}{ Car mod }
&\rotatebox[origin=c]{90}{ Car hard }
& \rotatebox[origin=c]{90}{ Ped. easy }
& \rotatebox[origin=c]{90}{ Ped. mod }
& \rotatebox[origin=c]{90}{ Ped. hard }
& \rotatebox[origin=c]{90}{ Cyclist easy }
& \rotatebox[origin=c]{90}{ Cyclist mod }
&\rotatebox[origin=c]{90}{ Cyclist hard}\\
\hline
Low res & 78.8 & 65.0 & 59.7 & 7.5 & 4.6 & 4.2 & 22.9 & 16.2 & 14.9\\
High res & 87.7 & 88.0 & 86.6 & 57.3 & 51.4 & 46.8 & 81.5 & 62.9 & 59.0 \\
Upsampled & 76.2 & 57.1 & 52.4 & 27.2 & 21.2 & 18.7 & 21.9 & 13.0 & 11.7 \\
\hline
\end{tabular}}    
\end{center}
\caption{Results of the PointPillar for KITTI dataset trained on high resolution lidar data and validated on decimated, original and upsampled by our method \textit{val} sets.}
\label{tab:detection}
\end{table}

To perform a quantitative evaluation of the results, we assess the generated data using traditional metrics such as CD, Hausdorf distance (HD), EMD, and introduced SWD. For the best-performing network snapshots, the metrics mentioned above are reported in table~\ref{tab:results}. Our generated data reveals substantial performance improvement concerning HD and SWD.

Furthermore, to verify our hypothesis regarding the SWD loss, we study the effects of different losses CD, EMD, and SWD on the upsampling performance of our model. Here we train our network on the KITTI dataset with the patches of size $1024$ and upsampling ratio $2$ with a similar setup as described before but with varying loss functions. The results of this study are demonstrated in Fig.~\ref{fig:ablation_study}. They confirm our initial assumption and are aligned with the observations made before. Similarly to the reconstruction experiment here, CD leads to sparseness areas, and EMD leads to distortions. That observation also confirms the findings of previous works, which state that CD is a weaker metric than 1-Wasserstein and SWD approximation is more accurate than approximated EMD.

\section{Conclusion}
In this work, we provide a network for lidar point cloud upsampling, which aims to reconstruct the fine-grained lidar scan patterns accurately. Our method reveals the required sensitivity to lidar scan lines and shows convincing results on lidar upsampling on a smaller scale of scan patches. Our deep network utilizes the edge-aware dense convolutions for feature extraction and expansion together with Sliced Wasserstein Distance. Both key features enable our method to perform lidar upsampling in a direct, lightweight one-stage fashion, excluding the coarse and fine reconstruction. The proposed method improves on both qualitatively and quantitatively measured baseline methods with HD, EMD, and SWD metrics. One of the limitations of our approach, though, is the scale of processed samples. The extension of our method to large-scale lidar scans ($\geq 100K$ points) remains for future work.

\section{Acknowledgment}
The research leading to these results is funded by the German Federal Ministry for Economic Affairs and Energy within the project KI Delta Learning. The authors would like to thank the consortium for the successful cooperation.

{
\bibliographystyle{ieee}
\bibliography{literature}

\begin{thebibliography}{10}\itemsep=-1pt

\bibitem{Achlioptas2018}
P.~Achlioptas, O.~Diamanti, I.~Mitliagkas, and L.~J. Guibas.
\newblock Learning representations and generative models for 3d point clouds.
\newblock In {\em ICML}, 2018.

\bibitem{Behley2019}
J.~Behley, M.~Garbade, A.~Milioto, J.~Quenzel, S.~Behnke, C.~Stachniss, and
  J.~Gall.
\newblock {SemanticKITTI: A Dataset for Semantic Scene Understanding of LiDAR
  Sequences}.
\newblock In {\em ICCV}, 2019.

\bibitem{Besic2022}
B.~Besic, N.~Gosala, D.~Cattaneo, and A.~Valada.
\newblock Unsupervised domain adaptation for lidar panoptic segmentation.
\newblock In {\em RAL}, 2022.

\bibitem{Caccia2019}
L.~Caccia, H.~van Hoof, A.~Courville, and J.~Pineau.
\newblock Deep generative modeling of lidar data.
\newblock In {\em IROS}, 2019.

\bibitem{Cai2020}
R.~Cai, G.~Yang, H.~Averbuch-Elor, Z.~Hao, S.~Belongie, N.~Snavely, and
  B.~Hariharan.
\newblock Learning gradient fields for shape generation.
\newblock In {\em ECCV}, 2020.

\bibitem{Chang2015}
A.~X. Chang, T.~Funkhouser, L.~Guibas, P.~Hanrahan, Q.~Huang, Z.~Li,
  S.~Savarese, M.~Savva, S.~Song, H.~Su, J.~Xiao, L.~Yi, and F.~Yu.
\newblock {ShapeNet: An Information-Rich 3D Model Repository}.
\newblock In {\em arxiv}, 2015.

\bibitem{Coors2019}
B.~Coors, A.~P. Condurache, and A.~Geiger.
\newblock Nova: Learning to see in novel viewpoints and domains.
\newblock In {\em 3DV}, 2019.

\bibitem{Cuturi2013}
M.~Cuturi.
\newblock Sinkhorn distances: Lightspeed computation of optimal transportation
  distances.
\newblock In {\em NIPS}, 2013.

\bibitem{Deshpande2019}
I.~Deshpande, Y.-T. Hu, R.~Sun, A.~Pyrros, N.~Siddiqui, S.~Koyejo, Z.~Zhao,
  D.~Forsyth, and A.~Schwing.
\newblock Max-sliced wasserstein distance and its use for gans.
\newblock In {\em CVPR}, 2019.

\bibitem{Waymo2020}
S.~et~al.
\newblock Scalability in perception for autonomous driving: Waymo open dataset.
\newblock In {\em CVPR}, 2020.

\bibitem{Fan2017}
H.~Fan, H.~Su, and L.~Guibas.
\newblock A point set generation network for 3d object reconstruction from a
  single image.
\newblock In {\em CVPR}, 2017.

\bibitem{Geiger2012}
A.~Geiger, P.~Lenz, and R.~Urtasun.
\newblock Are we ready for autonomous driving? the kitti vision benchmark
  suite.
\newblock In {\em CVPR}, 2012.

\bibitem{Groueix2018}
T.~Groueix, M.~Fisher, V.~G. Kim, B.~Russell, and M.~Aubry.
\newblock Atlasnet: A papier-m\^ach\'e approach to learning 3d surface
  generation.
\newblock In {\em CVPR}, 2018.

\bibitem{Jung2022}
Y.~Jung, S.-W. Seo, and S.-W. Kim.
\newblock Fast point clouds upsampling with uncertainty quantification for
  autonomous vehicles.
\newblock In {\em ICRA}, 2022.

\bibitem{Kolouri2019}
S.~Kolouri, P.~E. Pope, C.~E. Martin, and G.~K. Rohde.
\newblock Sliced-wasserstein auto-encoders.
\newblock In {\em ICLR}, 2017.

\bibitem{Kwon2022}
Y.~Kwon, Youngsun adnd~Sung and S.-E. Yoon.
\newblock Implicit lidar network:lidar super-resolution via interpolation
  weight prediction.
\newblock In {\em ICRA}, 2022.

\bibitem{Li2018}
C.-L. Li, M.~Zaheer, Y.~Zhang, B.~Póczos, and R.~Salakhutdinov.
\newblock Point cloud gan.
\newblock In {\em ICLR}, 2018.

\bibitem{Li2019}
R.~Li, X.~Li, C.-W. Fu, D.~Cohen-Or, and P.-A. Heng.
\newblock Pu-gan: a point cloud upsampling adversarial network.
\newblock In {\em ICCV}, 2019.

\bibitem{Li2021}
R.~Li, X.~Li, P.-A. Heng, and C.-W. Fu.
\newblock Point cloud upsampling via disentangled refinement.
\newblock In {\em CVPR}, 2021.

\bibitem{Liu2020}
M.~Liu, L.~Sheng, S.~Yang, J.~Shao, and S.-M. Hu.
\newblock Morphing and sampling network for dense point cloud completion.
\newblock In {\em AAAI}, 2020.

\bibitem{Luo2021}
S.~Luo and W.~Hu.
\newblock Diffusion probabilistic models for 3d point cloud generation.
\newblock In {\em CVPR}, 2021.

\bibitem{Nguyen2021}
K.~Nguyen, N.~Ho, T.~Pham, and H.~Bui.
\newblock Distributional sliced-wasserstein and applications to generative
  modeling.
\newblock In {\em ICLR}, 2021.

\bibitem{NguyenTrung2021}
T.~Nguyen, Q.-H. Pham, T.~Le, T.~Pham, N.~Ho, and B.-S. Hua.
\newblock Point-set distances for learning representations of 3d point clouds.
\newblock In {\em ICCV}, 2021.

\bibitem{Pan2020}
L.~Pan.
\newblock Ecg: Edge-aware point cloud completion with graph convolution.
\newblock In {\em IEEE IROS}, 2020.

\bibitem{Qi2016}
C.~R. Qi, H.~Su, K.~Mo, and L.~J. Guibas.
\newblock Pointnet: Deep learning on point sets for 3d classification and
  segmentation.
\newblock In {\em CVPR}, 2017.

\bibitem{Qi2017}
C.~R. Qi, L.~Yi, H.~Su, and L.~J. Guibas.
\newblock Pointnet++: Deep hierarchical feature learning on point sets in a
  metric space.
\newblock In {\em NIPS}, 2017.

\bibitem{Qian2021}
G.~Qian, A.~Abualshour, G.~Li, A.~Thabet, and B.~Ghanem.
\newblock Pu-gcn: Point cloud upsampling using graph convolutional networks.
\newblock In {\em CVPR}, 2021.

\bibitem{Shan2020}
T.~Shan, J.~Wang, F.~Chen, P.~Szenher, and B.~Englot.
\newblock Simulation-based lidar super-resolution for ground vehicles.
\newblock In {\em RAS}, 2020.

\bibitem{Shi2019}
S.~Shi, X.~Wang, and H.~Li.
\newblock Pointrcnn: 3d object proposal generation and detection from point
  cloud.
\newblock In {\em CVPR}, 2019.

\bibitem{Shu2019}
D.~W. Shu, S.~W. Park, and J.~Kwon.
\newblock 3d point cloud generative adversarial network based on tree
  structured graph convolutions.
\newblock In {\em ICCV}, 2019.

\bibitem{Sugiyama2012}
M.~Sugiyama and M.~Kawanabe.
\newblock {\em Machine Learning in Non-Stationary Environments: Introduction to
  Covariate Shift Adaptation}.
\newblock The MIT Press, 2012.

\bibitem{Sun2020}
Y.~Sun, Y.~Wang, Z.~Liu, J.~E. Siegel, and S.~E. Sarma.
\newblock Pointgrow: Autoregressively learned point cloud generation with
  self-attention.
\newblock In {\em WACV}, 2020.

\bibitem{Tchapmi2019}
L.~P. Tchapmi, V.~Kosaraju, S.~H. Rezatofighi, I.~Reid, and S.~Savarese.
\newblock Topnet: Structural point cloud deco.
\newblock In {\em CVPR}, 2019.

\bibitem{Thomas2019}
H.~Thomas, C.~R. Qi, J.-E. Deschaud, B.~Marcotegui, F.~Goulette, and L.~J.
  Guibas.
\newblock Kpconv: Flexible and deformable convolution for point clouds.
\newblock In {\em ICCV}, 2019.

\bibitem{Valsesia2019}
D.~Valsesia, G.~Fracastoro, and E.~Magli.
\newblock Learning localized generative models for 3d point clouds via graph
  convolution.
\newblock In {\em ICLR}, 2019.

\bibitem{Vayer2020}
T.~Vayer, R.~Flamary, R.~Tavenard, L.~Chapel, and N.~Courty.
\newblock Sliced gromov-wasserstein.
\newblock In {\em NIPS}, 2020.

\bibitem{Wang2020}
X.~Wang, M.~H. Ang~Jr, and G.~H. Lee.
\newblock Cascaded refinement network for point cloud completion with
  self-supervision.
\newblock In {\em CVPR}, 2020.

\bibitem{Wang2019}
Y.~Wang, Y.~Sun, Z.~Liu, S.~E. Sarma, M.~M. Bronstein, and J.~M. Solomon.
\newblock Dynamic graph cnn for learning on point clouds.
\newblock In {\em Transactions on Graphics}, 2019.

\bibitem{Wu2019}
J.~Wu, Z.~Huang, D.~Acharya, W.~Li, J.~Thoma, D.~P. Paudel, and L.~Van~Gool.
\newblock Sliced wasserstein generative models.
\newblock In {\em CVPR}, 2019.

\bibitem{Yang2019}
G.~Yang, X.~Huang, Z.~Hao, M.-Y. Liu, S.~Belongie, and B.~Hariharan.
\newblock Pointflow: 3d point cloud generation with continuous normalizing
  flows.
\newblock In {\em ICCV}, 2019.

\bibitem{Yang2018}
Y.~Yang, C.~Feng, Y.~Shen, and D.~Tian.
\newblock Foldingnet: Point cloud auto-encoder via deep grid deformation.
\newblock In {\em CVPR}, 2018.

\bibitem{Yifan2019}
W.~Yifan, S.~Wu, H.~Huang, D.~Cohen-Or, and O.~Sorkine-Hornung.
\newblock Patch-based progressive 3d point set upsampling.
\newblock In {\em CVPR}, 2019.

\bibitem{Yin2021}
T.~Yin, X.~Zhou, and P.~Krähenbühl.
\newblock Center-based 3d object detection and tracking.
\newblock In {\em CVPR}, 2021.

\bibitem{Yu2018}
L.~Yu, X.~Li, C.-W. Fu, D.~Cohen-Or, and P.-A. Heng.
\newblock Ec-net: an edge-aware point set consolidation network.
\newblock In {\em ECCV}, 2018.

\bibitem{Yu2018PuNet}
L.~Yu, X.~Li, C.-W. Fu, and H.~P.-A. Cohen-Or, Daniel.
\newblock Pu-net: Point cloud upsampling network.
\newblock In {\em CVPR}, 2018.

\bibitem{Yuan2018}
W.~Yuan, T.~Khot, D.~Held, C.~Mertz, and M.~Hebert.
\newblock Pcn: Point completion network.
\newblock In {\em International Conference on 3D Vision}, 2018.

\bibitem{Zhang2018}
W.~Zhang, H.~Jiang, Z.~Yang, S.~Yamakawa, K.~Shimada, and L.~B. Kara.
\newblock Data-driven upsampling of point clouds.
\newblock In {\em CAD}, 2018.

\end{thebibliography}
}


\end{document}